\documentclass[a4paper]{article}

\usepackage{INTERSPEECH_v2}
\usepackage{times}
\usepackage{latexsym}

\usepackage{enumitem}

\usepackage{xcolor}
\usepackage{url}
\usepackage{array}
\usepackage{multirow}
\usepackage{amssymb}
\usepackage{amsmath}
\usepackage{mathtools}
\usepackage{amsthm}
\usepackage{xspace}
\usepackage{algorithm}
\usepackage{tikz-dependency}
\usepackage{cancel}
\usepackage{balance}

\usepackage{graphicx}
\usepackage{subfig}

\newcommand{\notes}[1]{}%{\it {\small {#1}}}}

\theoremstyle{definition}

\theoremstyle{plain}

\newcommand{\ith}[1]{\ensuremath{i^{{th}}}}

\newcount\permx
\newcount\permy
\def\permdot#1#2{
\permx=#1 \advance\permx by-1
\permy=#2 \advance\permy by-1
\psframe[fillcolor=black, fillstyle=solid]
(\permx,\permy)(#1, #2)
}

\newcommand{\boxnum}[1]{{\setlength{\fboxsep}{1pt}\raisebox{1pt}{\hspace{1pt}\fbox{\tiny #1}\hspace{1pt}}}}
\newcommand{\ind}[1]{\ensuremath{_{\kern-0.5pt\boxnum{#1}}}}

\def\namecite{\newcite}

\newcommand{\smallnt}[1]{\ensuremath{_{\mbox{\tiny PP}}}\xspace}

\newcommand{\pseudocode}{Algorithm}
\floatname{algorithm}{\pseudocode}

\iffalse

\else

\fi

\providecommand{\card}[1]{\lvert#1\rvert}  % cardinality |x|

\definecolor{c1}{gray}{0.45}
\definecolor{c2}{gray}{0.65}
\definecolor{c3}{gray}{0.8}

\newcommand{\mathcolorbox}[2]{\ensuremath\text{\colorbox{#1}{\ensuremath{#2\!}}}\xspace}

\title{Jointly Trained Sequential Labeling and Classification\\ by Sparse Attention Neural Networks}
\name{Mingbo Ma$^1$, Kai Zhao$^1$, Liang Huang$^1$, Bing Xiang$^2$, Bowen Zhou$^2$}
\address{
  $^1$ Department of EECS, Oregon State University, USA\\
  $^2$ IBM Watson Group, T. J. Watson Research Center, IBM, USA}
\email{\{mam,zhaok,liang.huang\}@oregonstate.edu, \{bingxia,zhou\}@us.ibm.com}

\begin{document}

\maketitle
\begin{abstract}
%\vspace{-0.2cm}
Sentence-level classification and sequential labeling are two fundamental tasks in language understanding. While these two tasks are usually modeled separately,
in reality, they are often correlated, for example in intent classification and slot filling, or in topic classification and named-entity recognition. 
In order to utilize the potential benefits from their correlations, 
we propose a jointly trained model for learning the two tasks simultaneously 
via Long Short-Term Memory (LSTM) networks. 
This model predicts the sentence-level category and the word-level label sequence from the stepwise output hidden representations of LSTM. 
We also introduce a novel mechanism of ``sparse attention'' to weigh words differently based on their semantic relevance to sentence-level classification. The proposed method outperforms baseline models on ATIS and TREC datasets.
\end{abstract}

%!TEX root = main.tex
\vspace{-0.2cm}
\section{Introduction}
\label{sec:intro}
%Natural Language Understanding (NLU) has drawn many research attention in the area of Natural Language Processing (NLP) for a few decades. 
We consider the dichotomy between two important tasks in spoken language understanding: the global task of sentence-level classification,
such as intention or sentiment, and the local task of 
sequence labeling or semantic constituent extraction, 
such as slot filling or named-entity recognitions (NER). %While sentence level understanding is usually treated as a standard multi-class classification problem,  semantic constituents extraction can be formulated as sequential labeling problem. 
Conventionally,
these two tasks are 
% modeled separately, with algorithms such as SVM \cite{Boser:1992} or logistic 
% regression for the former, and Conditional Random Field (CRF)\cite{Lafferty:2001} and structured perceptron \cite{collins:2002} for the latter.
modeled separately, with algorithms such as SVM \cite{Boser:1992} 
%or logistic regression 
for the former, and Conditional Random Fields (CRF)~\cite{Lafferty:2001} 
or structured perceptron \cite{collins:2002} for the latter.
% please cite svm, maxent, crf, and perceptron!

%There are many classical algorithms that can be used for sentence level classification, such as SVM or logistic regression. Conditional Random Field (CRF)\cite{Lafferty:2001} and maximum entropy Markov models \cite{McCallum:2000} are  used extensively for sequential labeling. 

In reality, however, these two tasks are often correlated.
%, and it is beneficial to model them jointly in some tasks. % since there is a great potential that these two tasks could benefit from each other. 
Consider the problems of sentence topic classification and NER in Figure~\ref{fig:apple}. %(or intent classification and slot-filling), 
Different sentence-level classifications provide different priors for each word's label;
for example if we know the sentence is about IT news 
then the word ``Apple" is almost certainly about the company.
Likewise, different word-level label sequence also 
influence the sentence-level category distribution;
for example if we know the word ``Apple" is about fruits 
then the sentence topic is more likely to be agricultural. 

Indeed, previous work has explored joint modeling between the two tasks.
For example,
%This intuition is confirmed by the success of the discrete CRF model for 
Jeong and Lee \cite{Jeong:2008} propose a discrete CRF model for 
joint training of sentence-level classification and sequence labeling. 
A follow-up work~\cite{puyang:2013}
leverages the feature representation power of convolution neural 
networks (CNNs) to make the CRF model generalize better for unseen data.
However, the above CRF-based methods still suffer from 
the following limitations:
firstly, although CRF is trained globally, 
it still lacks the ability for capturing long-term memory 
at each time step which is crucial for sentence-level classification 
and sequential labeling problem;
secondly, the CNNs-based CRF~\cite{puyang:2013} only use CNNs for nonlinear local feature
extraction. But globally, it is still a linear model which has limited generalization power to unseen data. 

\begin{figure}[!tbp]
	\centering	
	\begin{minipage}[b]{0.47\textwidth}
	\noindent \fbox{\parbox{1\textwidth}{%
		\textbf{Topic: \textit{IT news}}\\
		{\color{red} \textit{Apple}}\rq s new iPad will be out soon ...
		\hfill {\color{blue}(Company)}\\[0.5em]
		\textbf{Topic: \textit{Agriculture reports}}\\
		{\color{red} \textit{Apple}} harvest in Washington state .... \hfill{\color{blue}(Fruit)}\\[0.5em]
		\textbf{Topic: \textit{Tourist information}}\\
		``Big {\color{red} \textit{Apple}}'' is a nickname of NYC ...\hfill{\color{blue}(Location)}
		}}
        \vspace{-.3cm}
	\caption{\small The same word ``Apple'' has different meanings (see tags in blue) in different different topics.
	\label{fig:apple}}
  	\end{minipage}\\[0.4cm]
%  \vspace{1cm}
  \hfill
  	\begin{minipage}[b]{0.47\textwidth}
    \noindent \fbox{\parbox{1\textwidth}{%
		\textbf{Category: Geography}\\
		\colorbox{c1}{Texas} is located in the \colorbox{c2}{South} \colorbox{c2}{Central} region...\\[0.5em]
		\textbf{Category: Date or Time}\\
		The train leaves at \colorbox{c1}{nine} \colorbox{c2}{o'clock} \\[0.5em]
		% \textbf{Category: Medical}\\
		% People with \colorbox{c1}{Post-traumatic} \colorbox{c2}{Stress} \colorbox{c3}{Disorder}\\[0.5em]
		\textbf{Category: Positive review}\\
		A \colorbox{c1}{smart}, \colorbox{c1}{sweet} and \colorbox{c3}{playful} romantic comedy.
		}}
    \vspace{-.2cm}
	\caption{\small Examples of various magnitudes of attentions for sentence level classification. Darker words are more important.% (with higher weights).
	\label{fig:att}}
  	\end{minipage}
  	\vspace{-1cm}
\end{figure}

In order to overcome the two above issues, % by incorporating long-term memory 
%and nonlinearity (both in local and global), we propose a novel sequential 
we propose a novel LSTM-based model to jointly train sentence-level 
classification and sequence labeling,
which incorporates long-term memory 
and nonlinearity both locally and globally. We make the following contributions:
\begin{itemize}[leftmargin=.2cm]
 \setlength\itemsep{.1cm}
%which, to the best of our knowledge, 
%is the first LSTM-based model for joint training.
\item Our Long Short-Term Memory (LSTM)~\cite{Hochreiter:1997}
network analyzes the sentence using features extracted
by a convolutional layer.
Basically, each word-level label is greedily 
determined by the hidden representation from LSTM in each step,
and the global sentence category is determined by an aggregate (e.g., pooling) 
of hidden representations from all steps (Sec.~\ref{sec:model}).
%% Both sentence-level classification
%% and sequence labeling are solved using the hidden representations
%% from LSTM and will jointly guide the training of LSTM. 
\item We also propose a novel {\bf sparse attention model} which promotes important words in a given sentence 
%by allocating higher weights 
and demotes semantically vacuous words %(such as stop words) %(like ``an'' or ``the'') 
%by signing lower weights or even $0$ weights 
(Fig.~\ref{fig:att}; Sec.~\ref{sec:att}).
\item Finally, we develop a {\bf latent variable} version of our jointly trained model
which can be trained for the single task of 
%extend our joint model to 
sentence classification 
by treating word-level labels as latent information (Sec.~\ref{sec:latent}). % when sequence labels are unknown. 
% To the best of our knowledge, this is the first full nerual network based model that exceeds the performance of conventional joint learning models.
\end{itemize}

\section{LSTM for Labeling and Classification}
\label{sec:lstm_cnn}
\begin{figure*}[!ht]
\centering
\includegraphics[width=14cm,height=4.5cm]{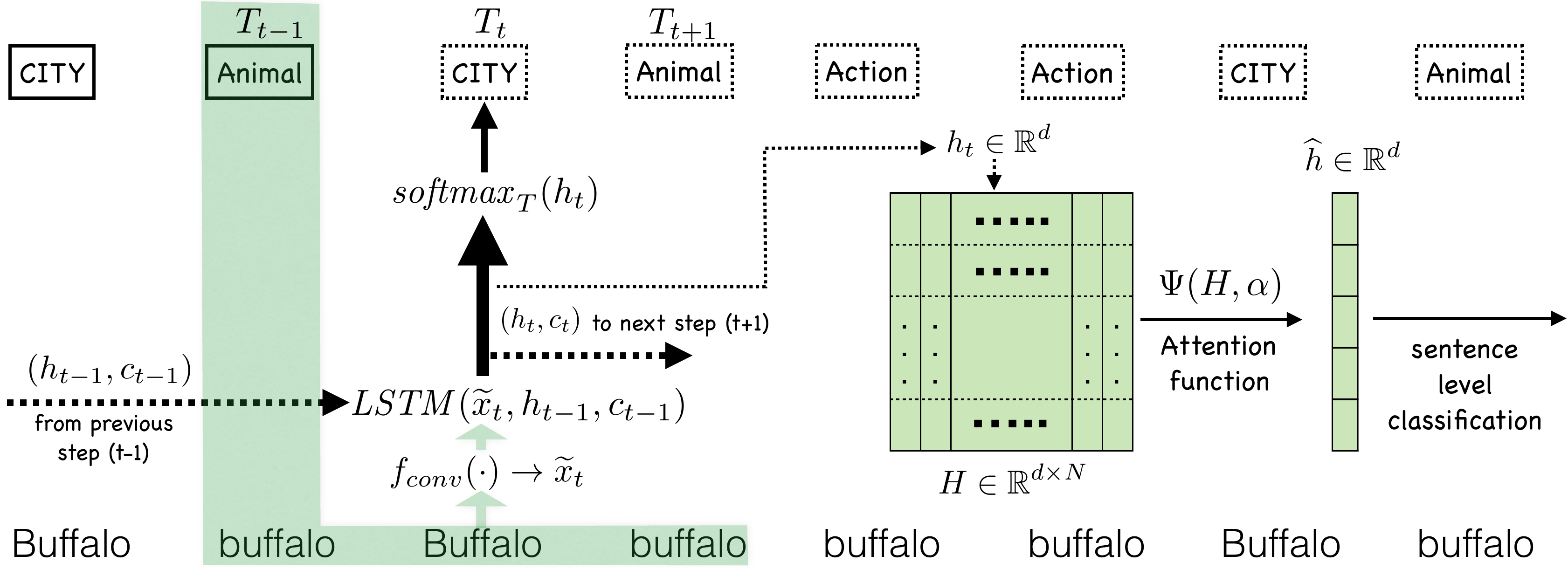}
\caption{\small Jointly trained sequence labeling and sentence classification. The green mask means that convolution operate between one previous tag and two surrounding words (when window size is 3). $\Psi$ is the attention function in Eq.~\ref{eq:att}. Here the sentence length is $N$=8.}%More details can be found in Sec.~\ref{sec:model}}
\label{fig:model}
\vspace{-5mm}
\end{figure*}

% Here we first briefly review the popular LSTM and CNN models %in deep learning. %LSTM is preferred for sequential labeling and CNN is widely used in sentence level classification.
% for sentence modeling
% to establish the mathematical notations. 
% %They are the backbones of our joint model.
\vspace{-2mm}
% \subsection{Long Short-Term Memory}
We use LSTM
to model the representation of the sentence at each word step,
which
is powerful in modeling sentence semantics \cite{sutskever2014sequence,cheng2016long}.
Assume the length of the sentence is $N$. 
LSTM represents the meaning of the sentence at the $t$-th word
%which is usually represented by its word embedding vector
%$x_t$, is represented 
by a pair of vectors $(h_t, c_t) \in (\mathbb{R}^d, \mathbb{R}^d)$,
where $h_t$ is the output hidden representation
of the word, $c_t$ is the memory of the network, 
and $d$ is the number of dimensions of the representation space:

\begin{equation}
\begin{bmatrix}
i_t\\
f_t\\
o_t\\
\hat{c}_t
\end{bmatrix}
=
\begin{bmatrix}
\sigma\\
\sigma\\
\sigma\\
\text{tanh}
\end{bmatrix}
W_{\text{LSTM}} \cdot [x_t, h_{t-1}]
\label{eq:lstm1}
\end{equation}
 \vspace{-1mm}
\begin{equation}
  c_t = f_t\odot c_{t-1} + i_t\odot \hat{c}_t
  \label{eq:lstm2}
\end{equation}
% \vspace{-1mm}
\begin{equation}
  h_t = o_t \odot \text{tanh}(c_t)
  \label{eq:lstm3}
\end{equation}
%The calculation of this representation pair is summarized by
%Eqs.~\ref{eq:lstm1}--\ref{eq:lstm3},
Here $x_t$ represents the $t$-th word,
which is usually the word embedding vector, 
and vectors $i_t, f_t$, and $o_t$ are gated activations
that control flow of hidden information.
The separation of the output representation $h_t$
from its internal memory $c_t$,
in principle, makes the knowledge about
the sentence prefix be remembered by 
the network for longer time
to interfere with the current output at word $x_t$.
These carefully designed activation gates
alleviate the problem of vanishing gradient problem
in vanilla recurrent neural network models.

\iffalse
Recurrent Neural Networks (RNNs) are powerful sequence modeling tools that can principally capture long term memories. However, the vanilla RNNs suffer from well-know vanishing or exploding gradient when the sequence get longer in practices \cite{Bengio:94}. In order to solve above issue, LSTM \cite{Hochreiter:1997} and Gated Recurrent Unit (GRU) \cite{Cho:2014} are proposed to control the gradient flow within the network. In this paper, we build our model based on LSTM, but it also can be adopted to GRU.

LSTM is applied

use lstm as global nonlinear model
\fi

In the task of sequence labeling, the label for 
each word is determined by hidden representation $h_t$.
As described in Eq.~\ref{eq:lstm3}, at time step $t$, we will get an output $h_t$ which represents all the current information. % in the form of feature vector. 
The probability distribution of the $t$-th word's label is calculated by
$\mathit{softmax}_T(W_T\cdot h_t)$, 
where weight $W_T\in \mathbb{R}^{d\times {\card{T}}}$
maps $h_t$ to the space of the labels,
and $\card{T}$ is the number of possible labels.
For the sequence labeling problem, 
the loss $\ell_{\text{seq}}$ is calculates as the sum of the 
Negative Log-Likelihood (NLL) over this label distribution 
$\mathit{softmax}_T(h_t)$ at each time step. 

In the task of sentence classification, the entire 
sentence representation is obtained by aggregating all history outputs $h_t$ which are stored in $H \in \mathbb{R}^{d \times N}$, where $N$ is the length of sequence.
Similar to CNNs, max pooling \cite{collobert+:2011,kim:2014,ma+:2015} operates over the history outputs $H$ to get the average activation $\hat{h} = \mathit{pooling} (H)$
% \begin{equation}
% \hat{h} = \mathit{pooling} (H),
% \label{eq:pooling}
% \vspace{-2mm}
% \end{equation}
 summarizing the entire outputs. 
Then, this sentence representation is passed to a fully connected soft-max layer
which outputs a distribution over sentence categories.

In above two different tasks, the hidden representation $h_t$ functions as a key component in different, separate ways.
In many cases, when sequence-level labels and sentence categories are both available we should use both 
information within the same framework by joint training
the two tasks.

% \subsection{Convolution Layer}
% \label{sec:cnn}

%LSTM processes a sequence of words by incrementally adding up 
%new content into memories. 

% \iffalse
% We formally define the chunk from the $i$-th word to the $(i+k)$-th word:
% \begin{equation}\label{eq:seq_con}
%     x_{i,k} = [x_i,x_{i+1}, \dots,x_{i+k} ]
% \end{equation}
% Sequential word concatenation $\vecx_{i,k}$ works as \ngram models which feeds contextual information into convolution operations. 
% For a given concatenated word sequence $\vecx_{i,k}$, 
% the convolution operation applies filter matrix $\mathbf{W}_{conv} \in  \vecR^{N \times kd}$ to $ \vecx_{i,k}\in R^{kd}$ with a bias term $b_{conv}$ described in equation \ref{eq:con_def}:
% \vspace{-2mm}
% \begin{equation}\label{eq:con_def}
%     \widetilde{x}_i = f_{conv}(\mathbf{W}_{conv} \cdot  \vecx_{i,k}+b_{conv} )
% \vspace{-2mm}
% \end{equation}
% where $f_{conv}$ is a non-linear activation function such as rectified linear unit (ReLu) or sigmoid function. Each row of $\mathbf{W}_{conv}$ represents one filter which measures the similarity between filter and input word chunk $x_{i,k}$. In order to capture enough variance of word combinations in each chunk, we usually apply $N$ different filters to the same chunk. Then $\widetilde{x}_i \in R^{N}$ is used as a contextual representation to replace $x_i \in R^d $ at each time step.  
% \fi

%!TEX root = main.tex

\section{Joint Sequence Classification \&  Labeling}
% In this section, we first describe our proposed joint model. Then, we introduce the idea of sparse attention. In the last part, we show our joint model could treat the label sequence as hidden variable when the ground truth of label sequence is not available.

\subsection{Joint Training Model}
\label{sec:model}

\begin{table*}[!ht]
\centering
\caption{\small Examples of ATIS sentences and annotated slots and categories.} 
\vspace{-2mm}%Sentence 1 is in Category ``Flight'' and sentence 2 is in Category ``Return''.}
\scalebox{0.8}{
\begin{tabular}{c| c c c c c c c c  c | c}
\hline
Sentence 1 & I & want & to & go & from & Denver & to & Boston & today  & Category\\
\hline
Slots   & O & O    & O  & O  & O    & B-FromCity & O & B-ToCity & B-Date & Flight \\   
\hline
\hline
Sentence 2 & to & come & back & to & Los & Angeles & on & Friday & evening & Category\\
\hline
Slots    & O  & O    & O    & O  & B-ReturnCity & I-ReturnCity & O & B-RETURN.DAY & B-RETURN.PERIODOFDAY & Return \\  
\end{tabular}
}
\label{tb:example}
\vspace{-4mm}
\end{table*}

%Fig.~\ref{fig:model} illustrates our model with a concrete running example. 
We aim at developing a model which could learn the label sequence and sentence-level category simultaneously.
To this end, we modify the standard LSTM structure to generate the word labels on the fly based on output $h_t$, and predict the sentence category with the sequence of $h_t$, $t=1,\dots, N$.

In LSTM, at time step $t$, only information of the current word $x_t$ is being fed into the network.
This mechanism overlooks the problem that the meanings of 
the same word in different contexts might vary ({\it cf}. Fig.~\ref{fig:apple}). In particular, words that follow $x_t$
are not represented at step $t$ in LSTM.

This observation motivates us to include more contextual information around the current word and previous tags as part of the input
to LSTM.
We employ convolutional neural network (CNN) to automatically mine
the meaningful knowledge from both the context of 
% the current
word $x_t$ and previous tags $T_{t-1}$, and use this knowledge as the input for LSTM.
%Fig.~\ref{fig:model} shows one classical NLP example that 
%a grammatically correct sentence only contains eight words ``buffalo''. 
%In order to capture better semantic meaning at each time step, contextual information should also be considered.
% \iffalse
% The original CNNs~\cite{LeCun95comparisonof}, applies convolution kernels on a series of continuous areas of given images, and was adapted to sentence modeling~\cite{collobert+:2011,kim:2014}. 
% Different from LSTM whose sentence presentation is added up cumulatively, convolution operation gets the feature representation at each time step by detecting certain words combinations in local chunks with associated filters. 
% Bigger chunk will capture more contextual information.
% \fi
We formally define the new input for LSTM as:
%\vspace{-2mm}
\begin{equation}
\vspace{-1mm}
\widetilde{x}_t = f_{\text{conv}}(W_{\text{conv}}\cdot x_{t,k}+b_{\text{conv}})
\label{eq:conv}
%\vspace{-1mm}
\end{equation}
where $f_{\text{conv}}$ is a non-linear activation function
like rectified linear unit (ReLU) or sigmoid,
and $x_{t,k}$ is a vector representing the context of word $x_t$ and previous tags $T_{t-k} ... T_{t-1}$,
e.g., the concatenation of the surrounding words and previous tags, $x_{t,k} = [x_{t-k}, \dots,x_{t+k}, \mathcolorbox{lightgray}{T_{t-k}, \dots, T_{t-1}}]$ in Eq.~\ref{eq:conv}, where the convolution window size is 
$2k+1$ and ${T_{t-1}}$ represents the embedding for tag $T_{t-1}$.
% \vspace{-2mm}
% \begin{equation}
% x_{t,k}=[x_{t-k},x_{t-k+1},\dots, x_{t+k}].
% \label{eq:conv_input1}
% \vspace{-1mm}
% \end{equation}
$W_{\text{conv}}$ is the collection of 
filters applied on the context.
During the convolution, each row of $W_{\text{conv}}$
is a filter that will be fired if it matches some 
useful pattern in the input context.
The convolutional layer functions as a feature extraction tool to 
learn meaningful representation from both words and tags automatically.
Note that the above contextual representation is different from bi-LSTM which only learns surrounding contextual of a given word. However, our model can learn both contextual and label information by convolution.

In our model, the joint training between sentence-level classification and sequential labeling is done in two directions: in forward pass, word label representation sequence is used for sentence level classification; during backward training, the sentence level prediction errors also fine-tune the label sequence.

Fig.~\ref{fig:model} illustrates our proposed model with 
one classical NLP exemplary sentence which 
only contains the word ``buffalo'' as the running example. 
At each time step, we first use the convolutional layer 
as a feature extractor to get the nonlinear feature combinations 
from the embeddings of words and tags. 
In the case of window size equaling to 1, the convolution operates over the $t-1$, $t$ and $t+1$ words and the $t-1$ tag. 
The contextual representation $x_{t, 1}=[x_{t-1}, x_t, x_{t+1}, T_{t-1}]$ is then fed into the convolution layer to find feature representation 
$\widetilde{x}_t$ following Eq.~\ref{eq:conv}.
$\widetilde{x}_t$ is used as the input for the following LSTM
to generate $h_t$ and $c_t$,
based on history information $h_{t-1}$ and 
previous cell information $c_{t-1}$. 
\iffalse
Sentence-level classification and sequential labeling depend  on the same 
meaning representation $h_t$:
\begin{itemize}
\item The first branch (left) uses $h_t$ as a 
latent representation to generate
 the tag $T_t$ through a softmax function $\mathit{softmax}_T$. 
 $T_t$ is saved for future local feature extraction 
 in next time step $t+1$. 
\item The second branch (upper right) saves latent representation $h_t$
 into a memory stack $H$ which save all the LSTM output through time. 
For sentence level classification,  $H$ will be fed into %Eq.~\ref{eq:att} to generate the sentence representation $\hat{h}$. 
a max-over-time pooling to extract the most useful information,
represented as $\hat{h}$.
In the last step, $\hat{h}$ will be pass through another 
softmax function to predict sentence-level category.
\end{itemize}
\fi

The example in Fig~\ref{fig:model} 
is a grammatically correct sentence in English~\cite{william:2012}. 
The word ``buffalo'' has three different meanings: 
Buffalo, NY (city), bison (animal), or bully (action). 
It is hard for standard LSTM to differentiate the different meanings of ``buffalo''
in different time step since the $x_t$ is the same all the time.
% When we use the standard LSTM which simply treats the 
% vanilla word embedding 
% $x_t$ as input from time step $t$, 
% it is hard to differentiate the words ``buffalo'' 
% in different time step since the $x_t$ is the same all the time. 
However, in our case, instead of simply using word representation $x_t$ itself, 
we also consider the contextual information
with their tags through 
convolution from $x_{t,k}$ (Eq.~\ref{eq:conv}).
% We get the nonlinear feature combinations through 
% convolution from surrounding words and tags for $x_{t,k}$.%(Eq.~\ref{eq:seq_con2}).

\vspace{-1mm}
\subsection{Sparse Attention}
\vspace{-1mm}
\label{sec:att}

% In the previous section, we state that we can get the 
% sentence-level meaning representation
% $\hat{h}$ by a simple average pooling on $H$,
% which assigns every outputs $h$ the same weights. 
% Although there are other pooling strategies 
% to get the final representation 
% such as max-pooling or statics pooling, 
% all these pooling strategies, however, operate on time direction 
% so that different words in sequence are treated equally,
% which is not the case in many scenarios. 
In the previous section, the sentence-level representation $\hat{h}$ is obtained 
by a simple average pooling on $H$. 
This process assumes every words contribute to the sentence equally,
which is not the case in many scenarios. 
Fig.~\ref{fig:att} shows a few examples of 
different words with different magnitudes of attention 
in different sentence categories. 
In order to incorporate these differences into consideration, 
we further propose a novel sparse attention constraint 
for sentence level classification. 
The sparse constraint assigns bigger weights for 
important words and 
lower the weights or even totally ignores the less meaningful words such like 
``the'' or ``a''.
% The sparse attention is formulated as the following
The attention-based sentence-level representation is formulated as follows:
%\vspace{-5mm}
\begin{equation}\label{eq:att}
\vspace{-2mm}
\begin{split}
\widehat{h} = \Psi(H,\alpha) = \sum_{h_t\in H} \psi  (h_t \cdot \alpha)h_t
\end{split}
\end{equation}
where $\alpha \in \mathbb{R}^d$ is an attention measurement 
which decides the importances for different inputs 
based on their semantics $h_i$. 
This importance is calculated through a nonlinear function $\psi$ which can be sigmoid or ReLU
and we use sigmoid in our case.
% . In our experiment, we use sigmoid to bound the weight value between $0$ and $1$.

Sparse Autoencoders \cite{andrew_sparse,Alireza} show that getting sparse representations in the hidden layers can further improve the performance. 
In our model, we apply similar sparse constraints by 
first calculating the average attention over the 
training samples in the same batch:
\vspace{-2mm}
\begin{equation}\label{eq:rho_hat_j}
\vspace{-2mm}
    \hat{\rho}_t = \frac{1}{m}\sum_{i=1}^m \psi  (h_t^i \cdot \alpha)
\vspace{-1mm}
\end{equation}
where $m$ is the size of training batch,
and $h_t^i$ is the output of LSTM
at step $t$ for example~$i$.
% \footnote{In practice, since the lengths
% of sentences in one batch are different, we
% adapt $m$ to be the number of LSTM outputs at step $t$ over the batch.}
% The goal of sparse attention is to enforce the constraint, $\forall t, \hat{\rho}_t = \rho$,
% % \vspace{-2mm}
% % \begin{equation}\label{eq:sparse_goal}
% % \forall t, \hat{\rho}_t = \rho
% % \vspace{-2mm}
% % \end{equation}
% where $\rho$ is 
% % the sparsity parameter which enforce 
% the desired sparsity of the attention. 
% Typically $\rho$ is set to be a small value close to zero. 
% % In order to satisfy this constraint, the attention over sequence almost set to be close to $0$.
In order to keep the above attention within a fix budget,
similar to Sparse Autoencoders \cite{andrew_sparse}, 
we have an extra penalty term as follows:
% In order to achieve the above objective, 
% there is an extra penalty term in our optimization objective 
% which reconstructs the original input 
% with as few attention as possible. 
% The most commonly used penalty term \cite{andrew_sparse} is the following:
\vspace{-2mm}
%\begin{eq}\label{eq:KL_sparse}
\[
    \mathit{KL}(\rho||\hat{\rho}_t)= \rho\log\frac{\rho}{\hat{\rho}_t} + (1-\rho)\log\frac{1-\rho}{1-\hat{\rho}_t},
    \]
%\end{eq}
where $\rho$ is the desired sparsity of the attention. 
%We also abuse the mathematics in that we use $\rho$
%to represent a uniform distribution of probability $\rho$
%on each dimension.
This penalty term uses KL divergence to measure the difference between two distributions.
Then our new objective is defined as follows:
\vspace{-1mm}
\begin{equation}\label{eq:loss_sparse}
    \ell_{\text{sparse}}(\cdot) = \ell_{\text{seq}}(\cdot) + \ell_{\text{sent}}(\cdot) + \beta \sum_{t=1}^N \mathit{KL}(\rho||\hat{\rho}_t),
\vspace{-1mm}
\end{equation}
% where $N$ is the number of hidden units and $\ell(\cdot)$ is network total loss consisting of 
% the sequence labeling loss $\ell_{\text{seq}}$, 
% sentence classification loss $\ell_{\text{sent}}$, 
% and the above sparsity penalty.
% $\beta$ controls the weight of the sparsity penalty term. 
% Note that the term $\hat{\rho}_t$ is implicitly controlled by optimizing $\alpha$ and output $h_t$ (Eq.~\ref{eq:rho_hat_j}). 
where $N$ is the number of hidden units, $\ell_{\text{seq}}$ is the sequence labeling loss,
and $\ell_{\text{sent}}$ is the sentence classification loss.
$\beta$ controls the weight of the sparsity penalty term. 
Note that the term $\hat{\rho}_t$ is implicitly controlled by optimizing $\alpha$ and output $h_t$.

% % \vspace{-2mm}
% \begin{itemize}
% \item Firstly, sparse attention is trained within one sentence. The goal is to find the most meaningful words which contribute to better sentence representation.
% This is different from dual sequences attention models \cite{rocktaschel2015reasoning,wang2015learning}. 
% % \item secondly, most attention models get the word-to-word attention through a softmax layer which tries to encourage one word and penalize other words in sequence. 
% % Our sparse attention tolerates multiple words with high attention 
% % for better classification loss. 
% \item Secondly, most attention models are softmax-based which gives a distribution over word or words. Softmax-based model has an exclusive property which allows cross influence between the source-side words. However, in sparse attention, this cross influence is not always necessary. The sum of attentions with sparse constraints does not have to be 1.
% \item Thirdly, most softmax-based attention models work well for pair of sentences. However, for single sentence modeling, the attention for words becomes implicit. In some cases, the distribution from softmax becomes flat (see discussion in Sec.~\ref{sec:visexp}).
% \end{itemize}

\vspace{-0.2cm}
\subsection{Label Sequence as Latent Variable}
\label{sec:latent}
In practice, it is expensive to annotate the data with both sequential label and sequence category.
In many cases, the sequence labels are missing since it requires
significantly more efforts to annotate the labels word-by-word.
However, even without this sequence labeling information, 
it is still helpful if we could utilize the 
possible hidden labels for each words.

In our proposed model, we could consider 
the sentence-level classification task as the major learning objective
and treat the unknown sequence labels as latent information. 
%This is another advantage over CRF based model since there is no latent tag representation could be generated by CRF.
The only adaptation we need to make is to replace the $T_t$ (tag embedding) with $h_t$ (output at time step $t$) in $x_{t,k}$. In this case, we 
%treat the possible tag information as a latent representation.
exploit the latent meaning representation 
to further improve the feature extraction of the convolutional layer.

%!TEX root = main.tex
\vspace{-1mm}
\section{Experiments}
\vspace{-2mm}

\begin{table}[b]%[!htbp]
\centering
\vspace{-0.3cm}
\caption{\small {\bf Main results:} Our jointly trained models compared with
various independent (marked $^\dagger$) and exisiting joint models.
The ``Slot'' column shows the F1 score
of sequence labeling,
and ``intent''  shows the error rates 
for sentence classification.
\vspace{-0.2cm}} %$\ddagger$: our implementation.}
\hspace{-0.45cm}\scalebox{0.87}{
\begin{tabular}{|p{3.1cm}|p{3.34cm}|p{0.7cm}|p{.6cm}|}
\hline
 & Model & Slot & Intent\\ \hline
\multirow{2}{*}{Independent Model} 
 & CRF \cite{Jeong:2008}     & 90.67$^\dagger$ &  7.91$^\dagger$                   \\
 & CNN \cite{puyang:2013}    & 92.43$^\dagger$ &  6.65$^\dagger$                   \\
 \hline 
\multirow{2}{*}{CRF Joint Model} &  TriCRF \cite{Jeong:2008}     &  \multicolumn{2}{l|}{94.42 \hfill  6.93}  \\
 & CNN TriCRF \cite{puyang:2013}           &  \multicolumn{2}{l|}{95.42 \hfill 5.91}  \\
 \hline
 \hline
 %\multirow{5}{*}
Independent Model & Vanilla LSTM (baseline)          &   93.74$^\dagger$  &  7.21$^\dagger$              \\
 %\cline{2-4}
 \hline
 & + joint             &  \multicolumn{2}{l|}{95.54 \hfill 6.32}  \\ 
{Jointly Trained Model}
 &  + joint + CNN                    &  \multicolumn{2}{l|}{\textbf{97.35} \hfill 5.96} \\
 %\multirow{0}{*}
 (Secs.~\ref{sec:model} \& \ref{sec:att})
 &  + joint + CNN + attention             &  \multicolumn{2}{l|}{96.73\hfill 5.71} \\
 &  + joint + CNN + sparse att.      &  \multicolumn{2}{l|}{96.98 \hfill  \textbf{5.12}} \\
 \hline
 \hline
 %\multirow{3}{*}
Independent Model & Vanilla LSTM (baseline)             & \  - &  7.21$^\dagger$              \\ 
\hline
\multirow{2}{*}{Seq.~Label as Latent Var.}
 &  + joint + CNN                                 &  \multicolumn{2}{l|}{\ -  \qquad\;\;    6.43}\\ 
% \multirow{0}{*}
\multirow{2}{*}{(Sec.~\ref{sec:latent})}
 &  + joint + CNN + attention                     &  \multicolumn{2}{l|}{\ - \qquad\;\; 5.61} \\
 &  + joint + CNN + sparse att.              &  \multicolumn{2}{l|}{\ - \qquad\;\; {\bf 5.42}}\\
\hline
\end{tabular}
}
\label{tb:exp1}
\vspace{-0.5cm}
\end{table}

%%%%%%%%%%%%%%%%%%%%%%%%%%%%%%%%%%%%%%%%%%%%%%%%%%%%
\begin{figure*}[!ht]
\begin{minipage}{2.25in}
\scalebox{0.85}{
\begin{tabular}{ |l|r| }
\hline
 Model & Sent.~Acc.\!\!\! \\
 \hline
CNN non-static \cite{kim:2014} &  93.6 \\
\hline
CNN multichannel \cite{kim:2014} &  92.2 \\
\hline
Deep CNN \cite{blunsom:2014} &  93.0 \\
\hline
\hline
Independent LSTM (baseline)&  92.2 \\
\hline
latent LSTM + CNN &  92.6 \\
\hline
latent LSTM + CNN + attention &  93.4 \\
\hline
latent LSTM + CNN + sparse att. &  \textbf{94.0} \\
\hline
\end{tabular}
}    
\caption{\small Sentence-level accuracy of our latent-variable model on TREC,
compared with various neural network-based models. }
\label{tb:exp2}
\end{minipage}
\hspace{0.25cm}
\begin{minipage}{4.3in}
\subfloat{%
  \includegraphics[clip,width=1\columnwidth]{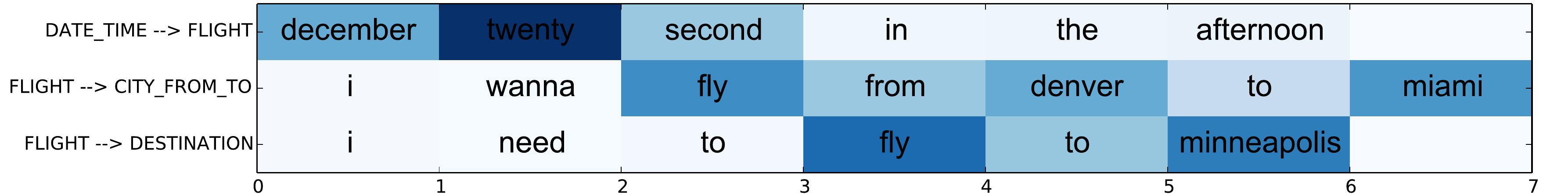}%
}\\
\subfloat{%
  \includegraphics[clip,width=1\columnwidth]{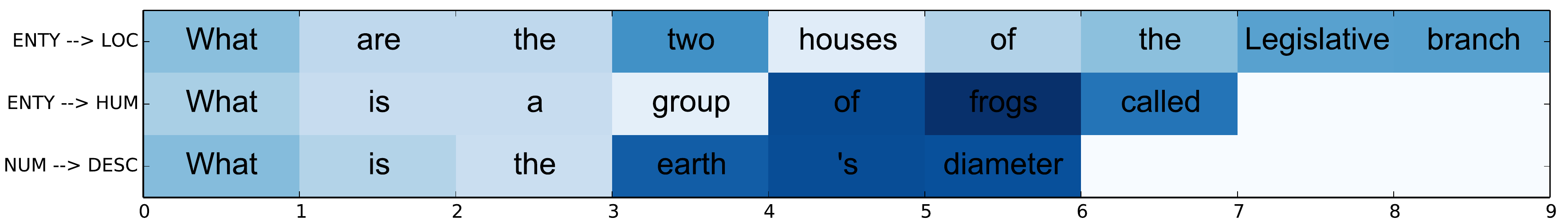}%
}
\caption{\small Examples that we outperform the model without sparse attention (LSTM+CNN). Higher weights are darker.}
\label{fig:vis_better}
\end{minipage}
\vspace{-0.4cm}
\end{figure*}
%%%%%%%%%%%%%%%%%%%%%%%%%%%%%%%%%%%%%%%%%%%%%%%%%%%%

%%%%%%%%%%%%%%%%%%%%%%%%%%%%%%%%%%%%%%%%%%%%%%%%%%%%
\begin{figure*}[!ht]
\centering
\includegraphics[width=0.7\textwidth]{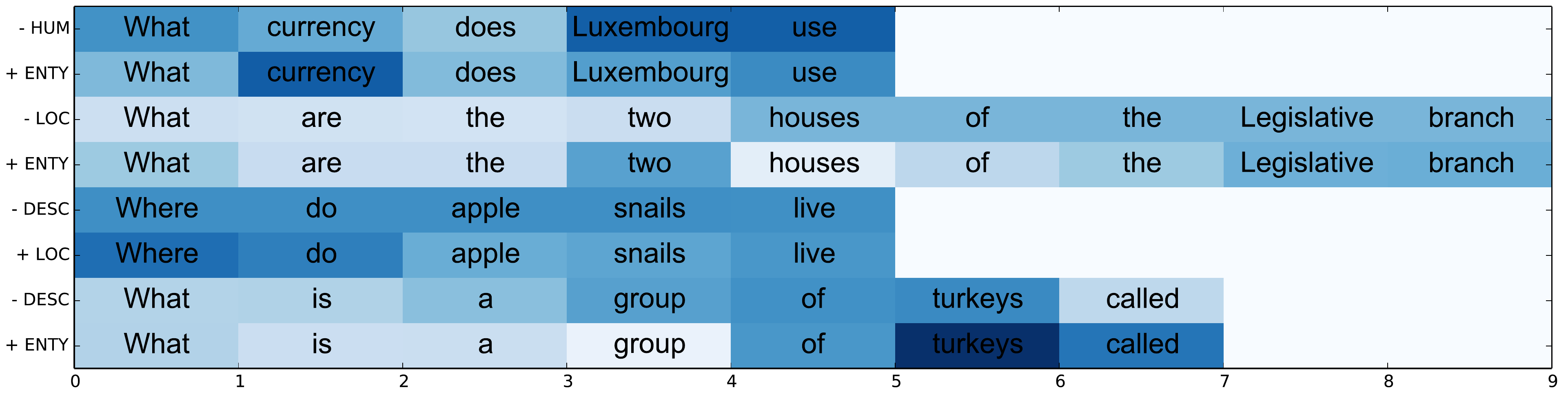}
\caption{\small Comparison between softmax-based attention (upper) and sparse attention (lower) for for some examples.
%the examples which sparse attention outperforms softmax-based attention. 
The sign `-' means mis-classified label, and `+' for the correct label. 
%For example, ``- HUM'' means the sentence is mis-classified as ``HUM'', likewise, ``+ ENTY'' means the sentence is corrected predicted as ``ENTY''. 
Darker blue represents higher weights.}
\vspace{-0.3cm}
\label{fig:sparse_softmax}
\vspace{-0.2cm}
\end{figure*}
%%%%%%%%%%%%%%%%%%%%%%%%%%%%%%%%%%%%%%%%%%%%%%

%In this section, we present a series of experiments to evaluate the performance of the proposed model. 

We start by evaluating the performance on a conventional joint learning task (Sec.~\ref{sec:jointexp}). 
Then we show the performance when we treat the label sequence as hidden information in Sec.~\ref{sec:latentexp}.
We also analyze some concrete examples to
show the performance of the sparse attention constraint in our model (Sec.~\ref{sec:visexp}). 

In the experiments, we set the convolution window sizes as 3, 5, and 7. There are 100 different filters for each window size. 
Word embeddings are randomly initialized in the ATIS experiments.
In the TREC experiments, we use $300$ dimension pre-trained word embeddings.
%Training batch size is 16. 
We use AdaDelta for optimization \cite{zeiler:2012}
with learning rate $0.001$ and minibatch 16. 
The weights in our framework are uniformly randomly initialized between $[-0.05, 0.05]$. 
We use ReLu at the convolutional layer and also regularize the feature with dropout 0.5 \cite{hinton:2014}.

We evaluate on ATIS\cite{Jeong:2008} and TREC \cite{Li:2002:LQC:1072228.1072378} datasets.
We follow the TriCRF paper\cite{Jeong:2008} in our evaluation on ATIS. 
There are $5,138$ dialogs
with 21 types of intents and 110 types of slots annotated
\footnote{\scriptsize{
Note we do not compare our performance with \cite{Guo:2014} since their ATIS dataset is not published and is different from our ATIS dataset.}}.
%We use the averaged results from 10-fold cross validation to demonstrate the final performance. 
% We use 10-fold cross validation in evaluation.
This dataset is first used for joint learning model with 
both slot and intent labels in the 
joint learning experiments (Sec.~\ref{sec:jointexp}). 
We later use it for evaluating the performance 
when the slot labels are not available (Sec.~\ref{sec:latentexp}).
% Two example sentences from the ATIS dataset are 
% shown in Table.~\ref{tb:example}.
%The performance of \cite{Guo:2014} does not exceed the performance of \cite{Jeong:2008} as shown in their experiments, so we only compare with \cite{Jeong:2008}.
The TREC 
dataset\footnote{\scriptsize{http://cogcomp.cs.illinois.edu/Data/QA/QC/}} 
is a factoid question classification dataset,
with $5,952$ sentences being classified 
into 6 categories.
%: abbreviation, entity, description, location and numeric.
%In the TREC dataset, the words in each sentence have very strong category 
%indicators such like entity or location. 
Since only the sentence-level categories are annotated,
we treat the unknown tags as latent information 
in our experiments (Sec.~\ref{sec:latentexp}). 

\vspace{-2mm}
\subsection{Joint Training Experiments}
\label{sec:jointexp}

We first perform the joint training experiments on the ATIS dataset.
Tab.~\ref{tb:example} shows two examples from the ATIS dataset.
As mentioned in Sec.~\ref{sec:att},
only a few keywords in the two sentences in Tab.~\ref{tb:example}
are relevant to determining the category,
i.e., locations (cities)
and date for sentence I, and locations, date, and
time for sentence II. Our model
should be able to recognize these important
keywords and predict the categories mostly based on them.
Once the model knows the tags of
the words in the sentence, it is straightforward
to determine the categories of the sentence.%,and vice versa.

% \begin{figure}[!ht]
% \begin{center}
% \noindent \fbox{\parbox{0.45\textwidth}{%
% \textbf{Category: Hotel}\\
% Yes I 'd like to \colorbox{c3}{stay} at the \colorbox{c1}{Hilton} \\[0.5em]
% \textbf{Category: Hotel}\\
% Do you \colorbox{c3}{have} a \colorbox{c1}{Marriott}.\\[0.5em]
% \textbf{Category: Return}\\
% And \colorbox{c1}{return} on \colorbox{c3}{December} \colorbox{c3}{fifteenth} in the \colorbox{c3}{afternoon} \\[0.5em]
% \textbf{Category: City}\\
% \colorbox{c1}{Portland} Oregon  \\[0.5em]
% \textbf{Category: Date or Time}\\
% \colorbox{c3}{In} the \colorbox{c1}{morning}
% }}
% \end{center}
% \caption{Examples of some visualization from corrected predicted sentence).
% \label{fig:vis}}
% \end{figure} 

% \begin{figure}
% \begin{center}
% \noindent \fbox{\parbox{0.45 \textwidth}{%
% \textbf{Category: Date or Time $\rightarrow$ Flight}\\
% That 's \colorbox{c2}{May} \colorbox{c1}{second} \\[0.5em]
% \textbf{Category: Date or Time $\rightarrow$ Flight}\\
% \colorbox{c3}{December} \colorbox{c1}{twenty} \colorbox{c1}{second} in the afternoon \\[0.5em]
% \textbf{Category: Date or Time $\rightarrow$ Flight}\\
% I would like to leave at \colorbox{c2}{nine} \colorbox{c2}{O'clock} \colorbox{c3}{in} \colorbox{c3}{the} \colorbox{c1}{evening} \\[0.5em]
% \textbf{Category: Flight $\rightarrow$ Destination}\\
% I need to \colorbox{c1}{fly} \colorbox{c1}{to} \colorbox{c2}{Newark}
% }}
% \end{center}
% \caption{Cases we better than the model without sparse attention constraints.
% \label{fig:vis_better}}
% \end{figure} 

We show the performance of our model
comparing with other individually trained or
jointly trained models in Tab.~\ref{tb:exp1}.
%First we observe
%that for discrete models like CRF,
%joint training does outperform
%individually trained model.
We observe that
due to its strong generalization power,
neural networks based models outperform
discrete models with an impressive margin.
Our jointly trained neural model
achieves the best performance,
with an F1 boost of $\sim$2 points in slot filling.
After adding the sparse attention constraint,
our model achieves the lowest error rate
for the sentence classification on ATIS.

\vspace{-0.2cm}
\subsection{Label Sequence as Latent Variable}
\label{sec:latentexp}
We further show the performance of our jointly trained model
when the sequence label information is missing.
%This task is more practical since 
%usually it is non-trivial to annotate the label
%for each word in the sentence, while
%annotating the sentence-level category is significantly easier.
%We first conduct the experiments on the ATIS dataset.
Tab.~\ref{tb:exp1} (bottom) shows the results on ATIS dataset.
There is a small increase of error rate
when sequence labels 
are unobserved, but our model
still outperforms existing models.
%due to the strong power of the convolutional
%layer in extracting useful features
%from the representations of the latent labels.
%Also, adding the sparse constraint
%lowers the error rates.
%We also evaluated our model on the TREC dataset,
%which is widely used for sentence category classification,
%where there is no sequence labels annotated.
Similarly, Fig.~\ref{tb:exp2} compares our latent-variable model
with conventional (non-latent) neural models on TREC dataset
(in sentence category accuracy).
Our model %still %achieves the best sentence-level accuracy
outperforms others
after adding sparse attention. %constraint.

\vspace{-0.2cm}
\subsection{Sparsity Visualization}
\label{sec:visexp}
% %%%%%%%%%%%%%%%%%%%%%%%%%%%%%%%%%%%%%%%%%%
% \begin{figure}[t]
% \subfloat{%
%   \includegraphics[clip,width=1\columnwidth]{figs/fig1_b2_2_b0.pdf}%
% }

% \subfloat{%
%   \includegraphics[clip,width=1\columnwidth]{figs/fig_b2_2_b0.pdf}%
% }
% \caption{\small Examples that we outperform the model without sparse attention (LSTM+CNN). Higher weights are darker.}
% \label{fig:vis_better}
% \vspace{-0.8cm}
% \end{figure}

% We show a few examples which we predict correctly the sentence level category in Fig.~\ref{fig:vis}. The first two examples are all about Hotel. And we get the most attentions which are focused on ``Hilton'' and ``Marriott''. In the third example, the attentions are not only focused on the word ``return'' but also focused on the date and time. The fourth example only have one word ``Portland'' that has attention. The word ``Oregon'' is totally ignored. 

In Fig.~\ref{fig:vis_better}, we compare the sparse attention model to the model without sparse constraints.
We list a few examples that the sparse attention is better than the one without sparse attention constraint. The labels on the right side are mis-classified by the model without sparse attention constraint. The label on the left side of the arrow is the ground truth. %The last example is the most interesting one. There are two words ``to'' in the sentence. Only the second ``to'' which are surrounded by ``fly'' and ``Newark'' are highlighted. The reason is that we take contextual information into consideration at each time step, and two different ``to'' get different representation from their contexts. 

In Fig.~\ref{fig:sparse_softmax}, we also compare the difference between two attention mechanisms: softmax-based attetion and sparse attention. From the first sentence, we can tell that softmax-based attention puts more emphasis on ``Luxembourg'' while sparse attention prefers ``currency'' which leads to the correct prediction of entity instead of human. In some cases, like the third example in Fig.~\ref{fig:sparse_softmax}, softmax-based sometimes gets confused by distributing the probabilities flatly. Compared with the attention model between dual sentences, the phenomenon of flat distribution is more obvious in single sentence attention.
Similar results can be found in the first figure in \cite{cheng:2016} as the word ``run'' is aligned to many unrelated words. It is possible that in single sentence attention, the softmax-based attention is easier to get confused since there is no obvious alignment or corresponding relationship between words for a given sentence or the words and their corresponding sentence category.

\vspace{-2mm}

%!TEX root = main.tex

\section{Related Works}
\label{sec:related}
\vspace{-2mm}

One neural network based model is proposed in~\cite{Guo:2014} for joint modeling the two tasks above with a parse-tree-based Recursive Neural Networks (RecNNs). 
However, as shown in their experiments, RecNNs fail to outperform most baselines. RecNNs-based jointly trained model is limited by two reasons: RecNNs's performance highly depends on the quality of parse-trees which are treated as inputs together with sentences; 
another problem of RecNNs is shared with all other Recurrent Neural Networks (RNN) based models which is hard to train due to gradient vanishing and exploding~\cite{Pascanu:2013}. 
When sentences get longer and parse-trees get deeper, RecNNs become harder to train. 

Our sparse attention model is different from other attention models 
in the following aspects:
Firstly, sparse attention is trained within one sentence. The goal is to find the most meaningful words which contribute to better sentence representation.
This is different from dual sequences attention models \cite{Bahdanau:2014,rocktaschel2015reasoning,wang2015learning}. 
Secondly, most attention models are softmax-based which gives a distribution over word or words. Softmax-based model has an exclusive property which allows cross influence between the source-side words. However, in sparse attention, this cross influence is not always necessary. The sum of attentions with sparse constraints does not have to be 1. 
Especially, our sparse attention is different from sparsemax attention~\cite{Andre:2016}.
Sparsemax tries to assign exactly
zero probability to some of its output variables,
but we try to control the sparsity of attention
by adjusting $\rho$ and $\beta$.
There are also neural-based efforts which only predict sequential labels,
e.g., \cite{Mesnil:2015,Kaisheng:2014}.

\vspace{-0.2cm}
%!TEX root = main.tex

\section{Conclusions}
\vspace{-1mm}
%In this paper 
We have presented a neural model that is jointly trained
on the two tasks of sentence classification and
sequence labeling, which %. We show that this joint learning
benefits from the correlation between the two tasks.
%% Our model employs LSTM to compute the semantic representation
%% for the words in the sentence, which is then trained to predict the sentence
%% category and sequence labels simultaneously.
%% Our convolution layer automatically distinguishes the
%% important features for the two tasks without 
%% human feature engineering. 
%% We further focus the attention of our model
%% on a small number of important
%% words, which improves the performance.
Our proposed models outperform both independent baselines
and existing joint models,
reaching the state-of-the-art in %some tasks.
either sentence classification or sequence labeling.

\section{Acknowledgment}
This work is supported in part by 
NSF IIS-1656051,
DARPA FA8750-13-2-0041 (DEFT),
DARPA N66001-17-2-4030 (XAI),
a Google Faculty Research Award,
and an HP Gift.

\newpage

\bibliographystyle{IEEEtran}

\bibliography{acl2016}

\end{document}